\documentclass{article}
\usepackage{spconf,amsmath,graphicx}

\title{Very Deep Multilingual Convolutional Neural Networks for LVCSR}
\name{Tom Sercu$^{1,2}$ \qquad Christian Puhrsch$^{1}$ \qquad Brian Kingsbury$^{2}$ \qquad Yann LeCun$^{1}$}
\address{$^{1}$ Center for Data Science, Courant Institute of Mathematical Sciences, New York University \\
\qquad   $^{2}$ IBM T. J. Watson Research Center, Yorktown Heights, NY, 10598, U.S.A. \\
\texttt{$^2$\{tsercu,bedk\}@us.ibm.com, $^1$cpuhrsch@nyu.edu,yann@cs.nyu.edu}
}
\begin{document}
\ninept
\maketitle
\begin{abstract}
Convolutional neural networks (CNNs) are a standard component of many current state-of-the-art
Large Vocabulary Continuous Speech Recognition (LVCSR) systems.
However, CNNs in LVCSR have not kept pace with recent advances in other domains where
deeper neural networks provide superior performance.
In this paper we propose a number of architectural advances in CNNs for LVCSR.
First, we introduce a very deep convolutional network architecture with up to 14 weight layers.
There are multiple convolutional layers before each pooling layer, 
with small 3$\times$3 kernels, inspired by the VGG Imagenet 2014 architecture.
Then, we introduce multilingual CNNs with multiple untied layers.
Finally, we introduce multi-scale input features aimed at exploiting more context at negligible computational cost.
We evaluate the improvements first on a Babel task for low resource speech recognition, obtaining an absolute
5.77\% WER improvement over the baseline PLP DNN by training our CNN on the combined data of six different languages.
We then evaluate the very deep CNNs on the Hub5'00 benchmark (using the 262 hours of SWB-1 training data) achieving
a word error rate of 11.8\% after cross-entropy training, 
a 1.4\% WER improvement (10.6\% relative) over the best published CNN result so far.
\end{abstract}
\begin{keywords}
Convolutional Networks, Multilingual, Acoustic Modeling, Speech Recognition, Neural Networks
\end{keywords}
\section{INTRODUCTION}
\label{sec:intro}

Convolutional Neural Networks (CNNs)~\cite{lecun1998gradient} have recently pushed the state of the art
on large-scale tasks in many domains dealing with natural data,
most notably in computer vision tasks like image classification~\cite{krizhevsky2012imagenet, simonyan2014very},
object detection \cite{sermanet2013pedestrian,girshick2014rich}, object localization \cite{sermanet2013overfeat}
and segmentation \cite{farabet2013learning}.

Early applications of neural nets to speech recognition used Time-Delay Neural Nets~\cite{waibel1989phoneme} which
can be seen as simple forms of CNNs without pooling or subsampling. Full-fledged CNNs with pooling and
subsampling were soon applied to speech recognition and combined with dynamic time warping~\cite{bottou1989experiments,bottou1990speaker}.
While the globally-trained combination of neural nets and HMMs for speech and handwriting goes back to the 1990s~\cite{bengio1992global,lecun1998gradient},
only due to recent developments \cite{seide2011conversational, mohamed2011deep, kingsbury2009lattice}
HMM/DNN hybrid modeling became dominant in ASR.
In the context of these hybrid models, the use of CNNs is relatively recent~\cite{abdel2012applying}.
CNNs were shown to achieve state of the art performance on the benchmark datasets
Broadcast News and Switchboard 300 \cite{sainath2013deep}.
However, in contrast to the trend in other domains where deeper architectures
are often shown to gain performance,
the classical CNN architecture in LVCSR \cite{sainath2013deep, soltau2014joint, saon2015ibm} 
has only two convolutional layers.

Our network architecture (Section \ref{ssec:deep}) is strongly inspired by the
work of Simonyan et al. \cite{simonyan2014very} (subsequently referred to as ``VGG Net'') which obtained
second place in the classification section of the Imagenet 2014 competition.
The central idea of VGG Net is to replace large convolutional kernels by a stack of 3$\times$3 kernels with ReLU nonlinearities
without pooling between these layers; 
The authors argue the advantage of this is twofold: (1) additional nonlinearity hence more expressive power, and (2) a reduced number of parameters.
Using these principles, very deep networks are trained with up to 19 weight layers (of which 16 are convolutional and 3 fully connected).
By contrast, the classical CNNs deployed in LVCSR have typically only two convolutional layers,
use large (9$\times$9) kernels in the first layer, and use sigmoid activation functions.
The first goal of this work is to adapt the VGG Net architecture to LVCSR.
Most closely related to this is \cite{bi2015very}, which also uses VGG Net-inspired CNNs for LVCSR
\footnote{This work was pursued independently of ours, and was published about two weeks before submission of this paper.}.
In contrast to our work, the architectures investigated in \cite{bi2015very} are quite different
and the paper only provides results from training on a non-standard Switchboard-51h dataset,
with WER not close to state of the art performance on Hub5'00.

In the context of low-resource language tasks, it can be crucial to leverage 
training data in languages other than the target language.
Therefore we trained multilingual deep CNNs, which we describe in Section \ref{ssec:ML}.
This is related to
multilingual neural networks in hybrid NN-HMM systems \cite{scanzio2008use}
which have been extended to multilingual bottleneck architectures for 
tandem models \cite{thomas2012multilingual, tuske2013investigation}
and have proven valuable for spoken term detection \cite{knill2013investigation}.
To our knowledge, no work has been published that extends the multilingual setup to CNNs.

The multi-scale features described in Section \ref{ssec:MS} aim at exploiting
more context at very low computational cost.
They are inspired by the recent success of combining information at multiple scales in tasks like 
traffic sign recognition \cite{sermanet2011traffic},
semantic segmentation \cite{farabet2013learning, long2014fully} and 
depth map prediction \cite{eigen2014depth}.
In LVCSR the multi-scale idea has been explored in tandem systems \cite{grezl2009investigation}
and the CLDNN architecture \cite{sainathconvolutional}.

\begin{table*}[ht]
\centering
\begin{tabular}{l | l | l | l | l | l }
\# Fmaps    & Classic \cite{sainath2013deep, soltau2014joint, saon2015ibm}  & VB(X)          & VC(X)          & VD(X)          & WD(X)          \\ \hline
64  &          & conv(3,64)     & conv(3,64)     & conv(3,64)     & conv(3,64)     \\
    &          & conv(64,64)    & conv(64,64)    & conv(64,64)    & conv(64,64)    \\ 
    &          & pool 1x3       & pool 1x2       & pool 1x2       & pool 1x2       \\ \hline
128 &          & conv(64, 128)  & conv(64, 128)  & conv(64, 128)  & conv(64, 128)  \\
    &          & conv(128, 128) & conv(128, 128) & conv(128, 128) & conv(128, 128) \\ 
    &          & pool 2x2       & pool 2x2       & pool 1x2       & pool 1x2       \\ \hline
256 &          &                & conv(128, 256) & conv(128, 256) & conv(128, 256) \\
    &          &                & conv(256, 256) & conv(256, 256) & conv(256, 256) \\
    &          &                &                &                & conv(256, 256) \\ 
    &          &                & pool 1x2       & pool 2x2       & pool 2x2       \\ \hline
512 & conv9x9(3,512) &          &                & conv(256, 512) & conv(256, 512) \\
    & pool 1x3  &       &                & conv(512, 512) & conv(512, 512) \\
    & conv3x4(512,512) &                &                &                & conv(512, 512) \\ 
    &          &                &                & pool 2x2       & pool 2x2       \\ \hline
    & \multicolumn{5}{c}{FC 2048}                                                  \\
    & \multicolumn{5}{c}{FC 2048}                                                  \\
    & \multicolumn{5}{c}{(FC 2048)}                                                \\
    & \multicolumn{5}{c}{FC output size}                                           \\ \hline
    & \multicolumn{5}{c}{Softmax}                                                  \\
\end{tabular}

\caption{\label{tab:deep}The configurations of our very deep CNNs for LVCSR. In all but the classic convnet,
    convolutional layers have 3$\times$3 kernels, thus kernel size is omitted. 
    The depth of the networks increases from left to right. 
    The deepest configuration, WDX, has 10 convolutional and 4 fully connected layers.
    The leftmost column indicates the number of output feature maps in each layer.
    The optional X means there are four fully connected layers instead of three (output layer included).
     }
\end{table*}
As training becomes more challenging with increasing depth,
we used two recently proposed optimization algorithms, Adadelta \cite{zeiler2012adadelta} and Adam \cite{kingma2014adam}
(Section \ref{ssec:training}).
Both algorithms are first order gradient-based optimization methods, which keep track of an estimate of
the first and second order moment of the gradient to tune the step size of each weight separately.

The rest of the paper is organized as follows.
In Section \ref{sec:novelties} we introduce the novel aspects of our work:
very deep CNN architectures in \ref{ssec:deep}, multilingual CNN training in \ref{ssec:ML},
multi-scale features in \ref{ssec:MS}, and training details in \ref{ssec:training}.
We then show experimental results on Babel in \ref{ssec:babel} and on Switchboard in \ref{ssec:swb}.

\section{ARCHITECTURAL AND TRAINING NOVELTIES}
\label{sec:novelties}

\subsection{Very Deep Convolutional Networks}
\label{ssec:deep}
The very deep convolutional networks we describe here are adaptations of the VGG Net
architecture \cite{simonyan2014very} to the LVCSR domain, where until now
networks with two convolutional layers dominated \cite{sainath2013deep, soltau2014joint, saon2015ibm}.
Table \ref{tab:deep} shows the configurations of the deep CNNs.
The deepest configuration, WDX, has 14 weight layers: 10 convolutional and 4 fully connected.
As in \cite{simonyan2014very}, we omit the Rectified Linear Unit (ReLU) layers following
every convolutional and fully connected layer.
The convolutional layers are written as conv(\{input feature maps\}--\{output feature maps\}) where 
each kernel is understood to be size 3$\times$3.
The pooling layers are written as (time x frequency) with stride equal to the pool size.

For architectures VDX and WDX, we apply zero padding of size 1 at every side before every convolution,
while for architecture VC(X) and VB(X) we use the convolutions to reduce the size of the feature maps, 
hence only in the higher layers of VC(X) padding is applied.

In contrast to \cite{simonyan2014very}, we do not reinitialize the deeper models with the shallower models.
Each model is trained from scratch with random initialization from a uniform distribution in the range 
$\lbrack -a, a \rbrack \text{~where~} a = (\text{kW} \times \text{kH} \times \text{numInputFeatureMaps})^{-\frac{1}{2}}$.
This follows the argument of \cite{glorot2010understanding} to initialize the weights such that
the variance of the activations on each layer does not explode or vanish during the forward pass.


\begin{figure}[htp]
    \centering
    \includegraphics[width=0.7\linewidth]{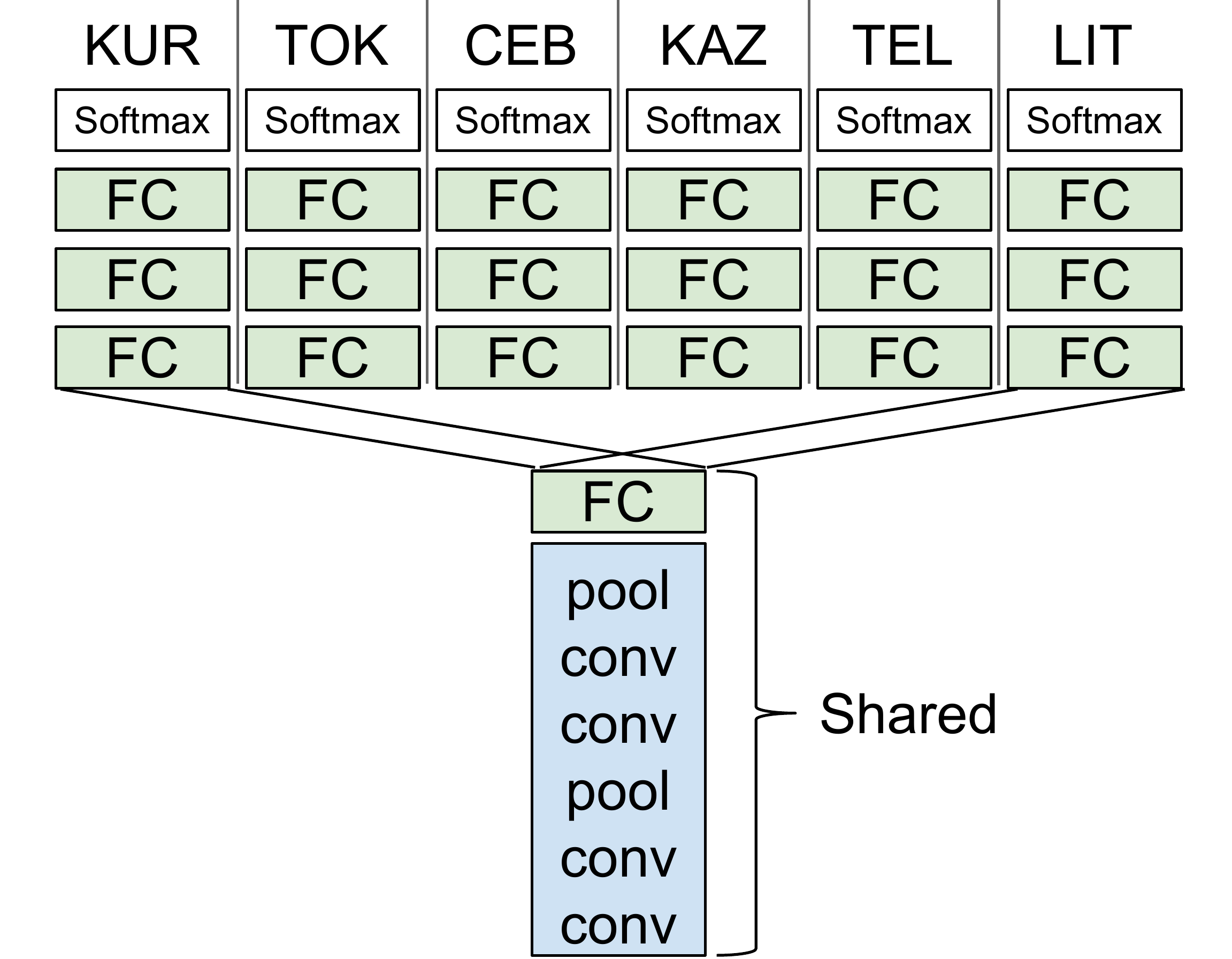}
    \caption{Multilingual VBX network with the last three layers untied. FC stands for Fully Connected layers.}
    \label{fig:multiling}
\end{figure}

\label{ssec:MS}
\begin{figure}[htp]
    \centering
    \includegraphics[width=0.7\linewidth]{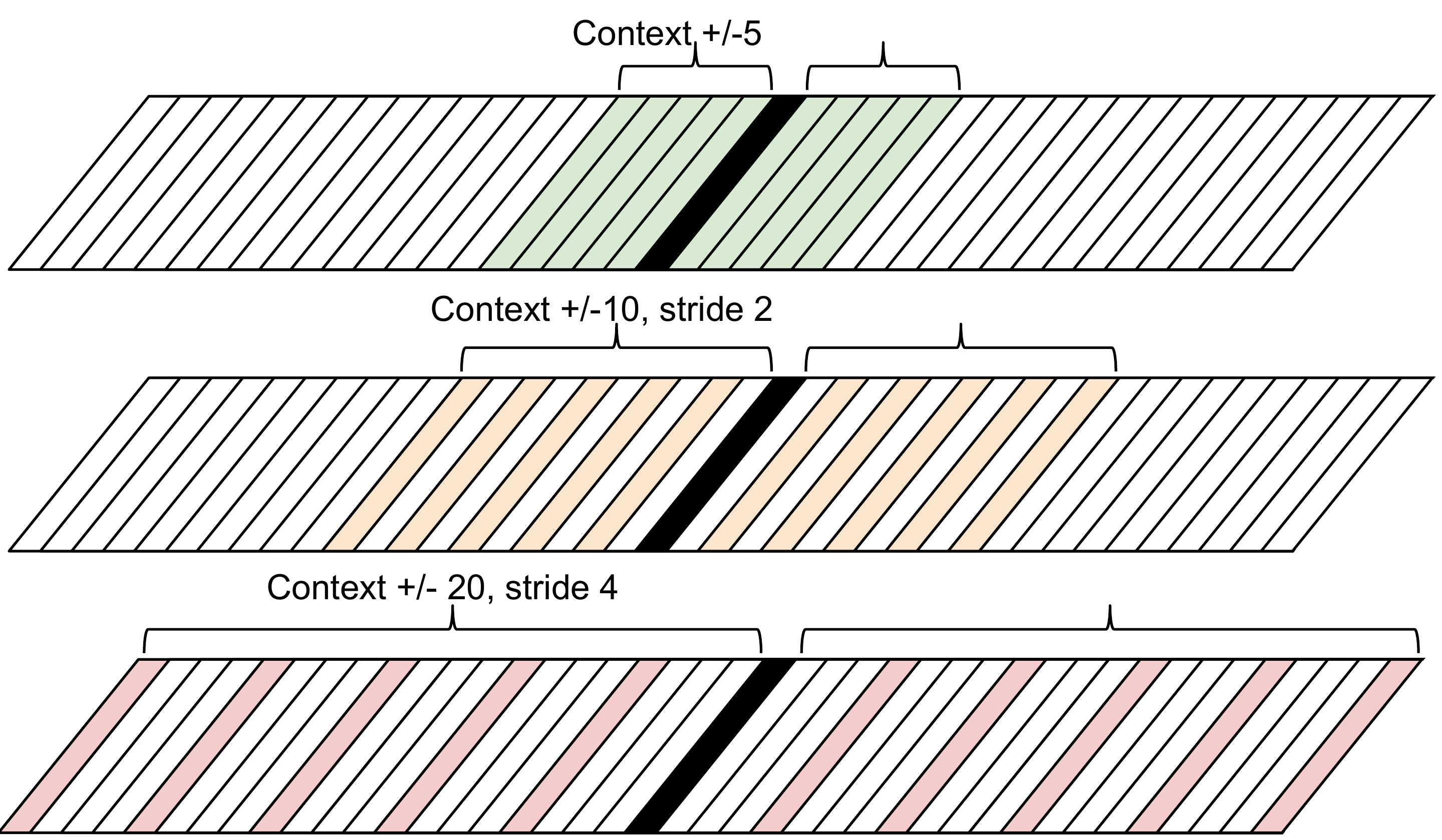}
    \caption{Multi-scale feature maps with context $\pm5$ and strides \{1,2,4\} (3S/5).
        The final size of each feature map along the time dimension is 11.
    The three 11$\times$40 input feature maps are stacked as input to the CNN, similar to how RGB channels form 
    3 input feature maps in an image.}
    \label{fig:MS_image}
\end{figure}
\subsection{Multilingual Convolutional Networks}
\label{ssec:ML}

Figure \ref{fig:multiling} shows a multilingual VBX network, which we used for
most of our Babel experiments.
It is similar to previous multilingual deep neural networks \cite{scanzio2008use}, 
with the main difference that the shared lower layers of the network are convolutional.

A second difference is that we untie more than only the last layer,
meaning that the weights and biases of multiple fully connected layers are
different for each language.
Since the output dimension of the convolutional stages is typically large when using 
large context windows, most of the weights are
in the first fully connected layer, which acts on the flattened output of the convolutional stages.
This is an argument to share this large, first fully connected layer across languages.
We experimentally confirmed that for all architectures,
untying all fully connected layers except the lowest one gives optimal performance,
with strong degradation if the first fully connected layer is also untied.
This untying corresponds to a view of the shared layers and the first fully connected layer as a shared
multilingual feature extractor, while the fully connected layers higher up form the classifier.

The multilingual CNN is trained in a round-robin fashion:
we process a mini-batch for each language before making an update to the weights.
In the shared part of the network the gradients of all mini-batches are accumulated between weight updates.
\subsection{Multi-scale feature maps}
The main goal of constructing multi-scale feature maps is to add more context 
without increasing the computational cost.
Figure \ref{fig:MS_image} illustrates the concept of multi-scale feature maps,
where additional input feature maps contain a larger view of the context of the frame
by downsampling larger context windows with different strides.
Kernels on the first convolutional layer are able to combine information from
multiple scales, i.e. different distances from the central frame.
Because the only difference for the convnet configuration is the first convolutional
layer having more feature maps, the additional computational cost
and number of parameters is small.

We found this style of multi-scale training to give small gains. 
Increasing the context size had a stronger positive impact, though
at the expense of increased computational cost.

\subsection{Training}
\label{ssec:training}

We use Adadelta \cite{zeiler2012adadelta} and Adam \cite{kingma2014adam} to do initial training
of the deep CNNs. Using Adadelta has two main advantages. Firstly, in our experience the optimization
problem converges much faster than with SGD; for the Babel experiments
we typically see convergence after about 40 million frames
using the 18 hours of Babel training data (after silence removal about 5.8 million frames).
Secondly, the optimal working point of Adadelta's hyperparameters $\epsilon$ and $\rho$ 
was stable across architectures, always giving optimal performance. This was crucial in order to 
explore architectural variations.
After initial training with Adadelta, we fine tune using SGD with a small learning rate.

Another aspect of training that improved our results is data balancing (something similar 
was done in \cite{sermanet2013overfeat}). We construct batches on the fly
by sampling target $y=CD_i$ with probability $p_i$, where $p_i$ is related to the frequency $f_i$ of
context dependent state $CD_i$ as $p_i = \frac{f_i^\gamma}{\sum_j f_j^\gamma}$.
After sampling $y$, we sample uniformly across all frames with that target.
The exponent $\gamma$ takes values between balanced training ($\gamma=0$) and
unbalanced training using the natural frequencies ($\gamma=1$).

In our experiments on Babel it proved optimal to start with $\gamma=0$ and raise
it during training to its final value of $\gamma=1$.
In our experiments on switchboard we varied $\gamma$ typically from 0.4 to 0.8
and decoded with HMM priors adjusted to match the final $p_i$ distribution.

\section{EXPERIMENTAL RESULTS}

\subsection{Babel}
\label{ssec:babel}

\begin{table}[ht]
\centering
\begin{tabular}{l | l | lllll}
         & DNN     & Classic & VB   & VBX  & VC   & VCX  \\ \hline
KUR      & 82.7 & 80.6    & 79.3 & 78.3 & 78   & 77.7 \\
TOK      & 62.6 & 59.4    & 57.1 & 56.1 & 54.3 & 54.5 \\
CEB      & 76.3 & 74.2    & 72.6 & 71.6 & 70.6 & 70.6 \\
KAZ      & 77.3 & 75.2    & 73.5 & 72.7 & 71   & 71.4 \\
TEL      & 87.0 & 85.4    & 83.7 & 83.7 & 82.4 & 82.7 \\
LIT      & 71.0 & 69.5    & 67.8 & 67.7 & 66   & 66.4 \\ \hline
IMPR     & 0.00 & 2.10    & 3.82 & 4.47 & \bf{5.77} & 5.60
\end{tabular}
\caption{\label{tab:depth}WER on Babel for different model architectures. Left to right is increasing depth. 
    The bottom row shows the absolute WER improvement over the CE PLP DNN baseline.
    Note that adding a fully connected layer for the 6-layer convolutional VC model (i.e. VCX) degrades performance.
}
\end{table}

\begin{table}[ht]
\centering
\begin{tabular}{l | l | lllll}
     & DNN  & 1L Clas    & 6L Clas    & 1L VC  & 6L VC  \\ \hline
KUR      & 82.7 & 82.8    & 80.6    & 81.3  & 78    \\
TOK      & 62.6 & 63.3    & 59.4    & 59.5  & 54.3  \\
CEB      & 76.3 & 76.7    & 74.2    & 73.2  & 70.6  \\
KAZ      & 77.3 & 77.7    & 75.2    & 74.4  & 71    \\
TEL      & 87.0 & 86.8    & 85.4    & 84.8  & 82.4  \\
LIT      & 71.0 & 72.7    & 69.5    & 69.8  & 66    \\ \hline
IMPR     & 0.00 & -0.52   & 2.10    & 2.32  & \bf{5.77}
\end{tabular}
\caption{\label{tab:multiling}WER on Babel for monolingual (1L, 3 hours of training data) versus 
    multilingual (6L, 18 hours of training data).
    When trained on a single language, the classical CNN architecture does slightly worse than the baseline DNN.
    However, the VC architecture gives an average 2.5 WER improvement even when trained on one language.
    As expected, for both models training multilingual gives a strong performance boost.}
\end{table}

\begin{table}[ht]
\centering
\begin{tabular}{l | l | lllll}
     & DNN  & 3S/20 & 1S/20 & 3S/8 & 1S/8 \\ \hline
KUR  & 82.7 & 78.1  & 78.4  & 78.4 & 79.2 \\
TOK  & 62.6 & 54.2  & 54.7  & 55.8 & 56.7 \\
CEB  & 76.3 & 70.3  & 70.4  & 71.6 & 71.8 \\
KAZ  & 77.3 & 71.1  & 71.8  & 72.5 & 72.8 \\
TEL  & 87.0 & 82.5  & 83.1  & 83.5 & 83.6 \\
LIT  & 71.0 & 66.2  & 67.3  & 66.9 & 67.5 \\ \hline
IMPR & 0.00 & \bf{5.75}  & 5.20  & 4.70 & 4.22
\end{tabular}
\caption{\label{tab:multiscale}WER for VC multi-scale training with different context windows.
    3S/20 stands for three scales with a context of $\pm$20.
    For 3S we use strides of 1, 2, and 4, while 1S just has stride 1, i.e. regular input features.
    Multi-scale features provide a modest gain. 
    Using larger context size gives a better improvement, however this comes
    at the cost of extra computation proportional to the context size in the convolutional layers.}
\end{table}

Our first set of experiments on Babel focuses on the multilingual and multi-scale
aspects of this work.
The IARPA Babel program is aimed at developing robust keyword search technology
for low resource languages.
Though the word error rates reported here are too high to be useful for simple speech to text applications,
useful keyword search (KWS) systems can still be built based on these ASR models.

As training data we use a combination of 6 languages, with 3 hours
of training data per language.
The languages used for training are languages from the second Option Period of the Babel
program, i.e. Kurmanji (KUR), Tok Pisin (TOK), Cebuano (CEB), Kazakh (KAZ), Telugu (TEL),
and Lithuanian (LIT).
The features used in these experiments are standard log-Mel features,
standardized with a global mean and variance shared across the speakers and langauges.
Unless explicitly mentioned, we use multi-scale features with context $\pm$20 in the Babel experiments.
We report results after cross-entropy training with adadelta ($\rho=0.985$, $\epsilon=1e{-10}$),
and $\gamma$ varying from 0 to 1.

We trained the multilingual deep CNN architecture on 6 Babel languages using
alignments from 6 baseline speaker independent HMM/DNN systems using PLP features,
with 1000 context dependent states.
The context dependent states are specific to each language.
Each baseline system is cross-entropy trained on a single language with 3 hours of data.
We will report the WER of the CNNs compared to the baseline DNN,
and summarize this in the average absolute WER improvement over the baseline DNN,
which gives one number to compare different models.
The WER improvements over the baseline DNN are fairly consistent across languages.

Tables \ref{tab:depth} through \ref{tab:multiscale}
show the results outlining the performance gains from the different architectural improvements 
discussed in Section \ref{ssec:deep}, \ref{ssec:ML}, and \ref{ssec:MS} respectively.
From table \ref{tab:multiling} note that even in the monolingual case
(3 hours of data) the VBX CNN architecture outperforms both the classical CNN and the baseline DNN.

\subsection{Switchboard 300}
\label{ssec:swb}
\begin{table}[ht]
\centering
\begin{tabular}{| l | l | l | l |}
    \hline
    & WER &  \# params (M) & \#M frames\\ \hline
    Classic 512 \cite{soltau2014joint}      & 13.2  & 41.2 & 1200 \\
    \hline
    Classic 256 ReLU (A+S)                        & 13.8  & 58.7 & 290 \\
    VCX (6 conv) (A+S)                            & 13.1  & 36.9 & 290 \\
    VDX (8 conv) (A+S)                            & 12.3  & 38.4 & 170 \\ 
    WDX (10 conv) (A+S)                           & 12.2  & 41.3 & 140 \\ 
    \hline
    VDX (8 conv) (S)                        & 11.9  & 38.4 & 340 \\  
    WDX (10 conv) (S)                       & \bf{11.8}  & 41.3 & 320 \\  
    \hline
\end{tabular}
\caption{\label{tab:hub5}Results on Hub5'00 SWB after training on the 262-hour SWB-1 dataset.
    We obtain 14.5\% relative improvement
    over our baseline adaptation of the classical CNN
    and 10.6\% relative improvement over \cite{soltau2014joint}.
    (A+S) means Adadelta + SGD finetuning. 
    (S) means the model was trained from random initialization using SGD.
    The last column gives the number of frames til convergence.
}
\end{table}

We evaluate our deep CNN architecture by training on the 262-hour SWB-1 training data,
and report the Word Error Rates on Hub5'00 SWB (table \ref{tab:hub5}).
The Switchboard experiments focus on the very deep aspect of our work.
Apart from not involving multilingual training,
we did not use multi-scale features in the Switchboard experiments, 
but did use speaker-dependent VTLN and deltas and double deltas as this is shown to help
performance for classical CNNs \cite{sainath2013deep}.

In the switchboard experiments, using a large context only gave marginal gains
which were not worth the computational cost,
so we worked with context windows of $\pm$8.
We use a data balancing value of $\gamma=0.8$, 
chosen from $[0.4,0.6,0.8,1.0]$.

After training with multiple combinations of Adam, Adadelta and SGD,
we settled on two possible strategies for optimization: the first strategy is to
use Adadelta or Adam for initial training,
followed by SGD finetuning. This way one can typically achieve good performance in minimal time.
The second strategy, training from scratch using only SGD, requires more training, 
however the performance is slightly superior.
Classical momentum yielded no gains and sometimes slight degradation over plain SGD.
We provide the results and total number of frames until convergence. 
Note that with the first strategy, we achieve 12.2\% WER
after 140M frames, i.e. only 1.5 passes through the dataset (which has 92.1M frames).
Using just SGD we achieve 11.8\% WER in 3.5 passes through the data.

We only present results after cross-entropy training, so we compare against
the best published cross-entropy trained CNNs.
The baseline is the work of Soltau et al. \cite{soltau2014joint} using
classical CNNs with 512 feature maps on both convolutional layers.
A second baseline is the work of Saon et al. \cite{saon2015ibm} which introduces
annealed dropout maxout CNN's with a large number of HMM states, achieving
12.6\% WER after cross-entropy training (not in the paper, from personal communication).
Note that these improvements could readily be integrated with our very deep CNN architectures.

\section{DISCUSSION}
In this paper we proposed a number of architectural advances in CNNs for LVCSR.
We introduced a very deep convolutional network architecture with small 3$\times$3 kernels
and multiple convolutional layers before each pooling layer, inspired by the VGG Imagenet 2014 architecture.
Our best performing model has 14 weight layers.
We also introduced multilingual CNNs which proved valuable in the context
of low resource speech recognition.
We introduced multi-scale input features aimed at exploiting more acoustic context
with minimal computational increase.
We showed an improvement of 2.50\% WER over a standard DNN PLP baseline using 3 hours of data,
and an improvement of 5.77\% WER by combining six languages to train on 18 hours of data.
We then showed results on Hub5'00 after training on 262 hours of SWB-1 data where we get 
11.8\% WER, which is
an improvement of 2.0\% WER (14.5\% relative) over our own baseline, and a 
1.4\% WER (10.6\% relative) improvement over the best
result published on classical CNNs after cross-entropy training \cite{soltau2014joint}.

We expect additional gains from sequence training, joint training with DNNs \cite{soltau2014joint}, and
integrating improvements like annealed dropout and maxout nonlinearities \cite{saon2015ibm}.

\section{ACKNOWLEDGEMENT}
This effort uses the very limited language packs from IARPA Babel
Program language collections IARPA-babel205b-v1.0a,
IARPA-babel207b-v1.0e, IARPA-babel301b-v2.0b, IARPA-babel302b-v1.0a,
IARPA-babel303b-v1.0a, and IARPA-babel304b-v1.0b.  This work is
supported by the Intelligence Advanced Research Projects Activity
(IARPA) via Department of Defense U.S. Army Research Laboratory
(DoD/ARL) contract number W911NF-12-C-0012. The U.S. Government is
authorized to reproduce and distribute reprints for Governmental
purposes notwithstanding any copyright annotation thereon.
Disclaimer: The views and conclusions contained herein are those of
the authors and should not be interpreted as necessarily representing
the official policies or endorsements, either expressed or implied, of
IARPA, DoD/ARL, or the U.S. Government.

We gratefully acknowledge the support of NVIDIA Corporation.

The authors would like to thank Pierre Sermanet for the initial code
base, George Saon, Vaibhava Goel, Etienne Marcheret and Xiaodong Cui for valuable
discussions and comments.


\bibliographystyle{IEEEbib}
\bibliography{refs}

\end{document}